\documentclass[10pt,twocolumn,letterpaper]{article}

\usepackage{cvpr}             

\definecolor{cvprblue}{rgb}{0.21,0.49,0.74}
\usepackage[pagebackref,breaklinks,colorlinks,allcolors=cvprblue]{hyperref}

\def\paperID{****} 
\def\confName{CVPR}
\def\confYear{2026}
\usepackage{amsmath}
\usepackage{amssymb}
\usepackage{xspace}
\usepackage[export]{adjustbox}
\usepackage{multirow}
\usepackage{threeparttable}
\usepackage{color, colortbl}
\usepackage{makecell}

\usepackage{xspace}
\usepackage[export]{adjustbox}
\usepackage{multirow}
\usepackage{threeparttable}
\usepackage{color, colortbl}
\usepackage{makecell}
\usepackage{algorithm2e}
\RestyleAlgo{ruled} 

\usepackage{xcolor} 
\usepackage{minted}
\usepackage[most]{tcolorbox}
\tcbuselibrary{skins, breakable}
\usepackage{newtxtext,newtxmath}

\definecolor{codebg}{rgb}{0.97,0.97,0.97}

\newcommand{\toolA}[0]{\includegraphics[scale=0.05,valign=c]{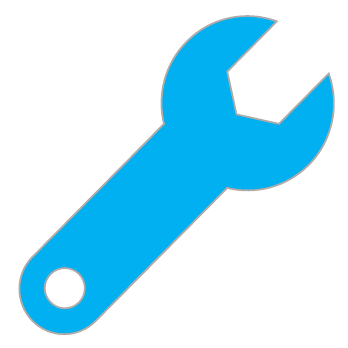}}
\newcommand{\toolB}[0]{\includegraphics[scale=0.05,valign=c]{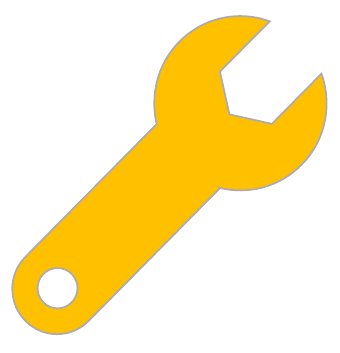}}

\newcommand{\memoryA}[0]{\includegraphics[scale=0.05,valign=c]{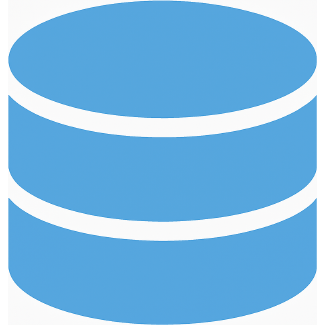}}
\newcommand{\memoryB}[0]{\includegraphics[scale=0.05,valign=c]{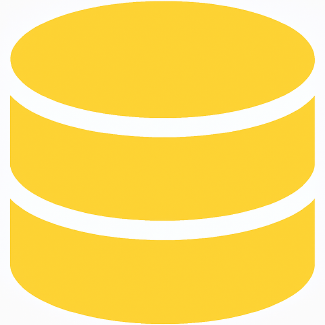}}
\newcommand{\memoryC}[0]{\includegraphics[scale=0.05,valign=c]{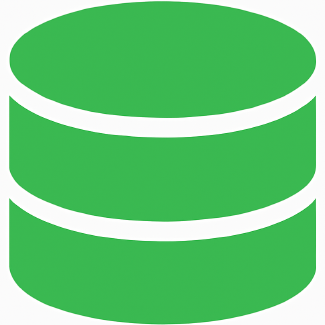}}

\title{VideoARM: Agentic Reasoning over Hierarchical Memory\\for Long-Form Video Understanding}

\author{
Yufei Yin\quad
Qianke Meng\quad
Minghao Chen\quad
Jiajun Ding\quad
Zhenwei Shao\quad
Zhou Yu\thanks{Zhou Yu is the corresponding author} \quad\\
\normalsize Key Laboratory of Complex Systems Modeling and Simulation\\
\normalsize School of Computer Science and Technology, Hangzhou Dianzi University, China \\
{\fontfamily{pcr}\selectfont \small \{yinyf, mqk, chenminghao, djj, shaozw, yuz\}@hdu.edu.cn}\\
\small Code: \url{https://milvlg.github.io/videoarm/}
}

\begin{document}
\maketitle
\begin{abstract}
Long-form video understanding remains challenging due to the extended temporal structure and dense multimodal cues. Despite recent progress, many existing approaches still rely on hand-crafted reasoning pipelines or employ token-consuming video preprocessing to guide MLLMs in autonomous reasoning. To overcome these limitations, we introduce VideoARM, an \underline{\textbf{A}}gentic \underline{\textbf{R}}easoning-over-hierarchical-\underline{\textbf{M}}emory paradigm for long-form video understanding. Instead of static, exhaustive preprocessing, VideoARM performs adaptive, on-the-fly agentic reasoning and memory construction. Specifically, VideoARM performs an adaptive and continuous loop of \textit{observing, thinking, acting, and memorizing}, where a controller autonomously invokes tools to interpret the video in a coarse-to-fine manner, thereby substantially reducing token consumption. In parallel, a hierarchical multimodal memory continuously captures and updates multi-level clues throughout the operation of the agent, providing precise contextual information to support the controller in decision-making. Experiments on prevalent benchmarks demonstrate that VideoARM outperforms the state-of-the-art method, DVD, while significantly reducing token consumption for long-form videos.
\end{abstract}

\section{Introduction}
\label{sec:intro}

Long-form video understanding~\cite{zhang2025deep,wang2024videoagent,fan2024videoagent,pang2025mr,wang2025videotree} is a fundamental yet challenging task for multimodal intelligence. Such videos, ranging from feature films and documentaries to lectures and meetings, contain densely structured information unfolding over extended time scales. Comprehending these sequences requires not only capturing fine-grained spatiotemporal details, but also reasoning about the dynamic evolution of long-range dependencies and semantic relationships throughout the narrative. Recent advances in multimodal large language models 
(MLLMs)~\cite{liu2023visual,openai2025o3,bai2025qwen2,damonlpsg2025videollama3,wang2025internvideo2} 
have shown strong potential, with capabilities in long-range reasoning, contextual association, and cross-modal alignment. Their ability to process long contexts and integrate multimodal cues provides a solid foundation for holistic long-form video understanding. 
\begin{figure}
    \centering
    \includegraphics[width=0.98\linewidth]{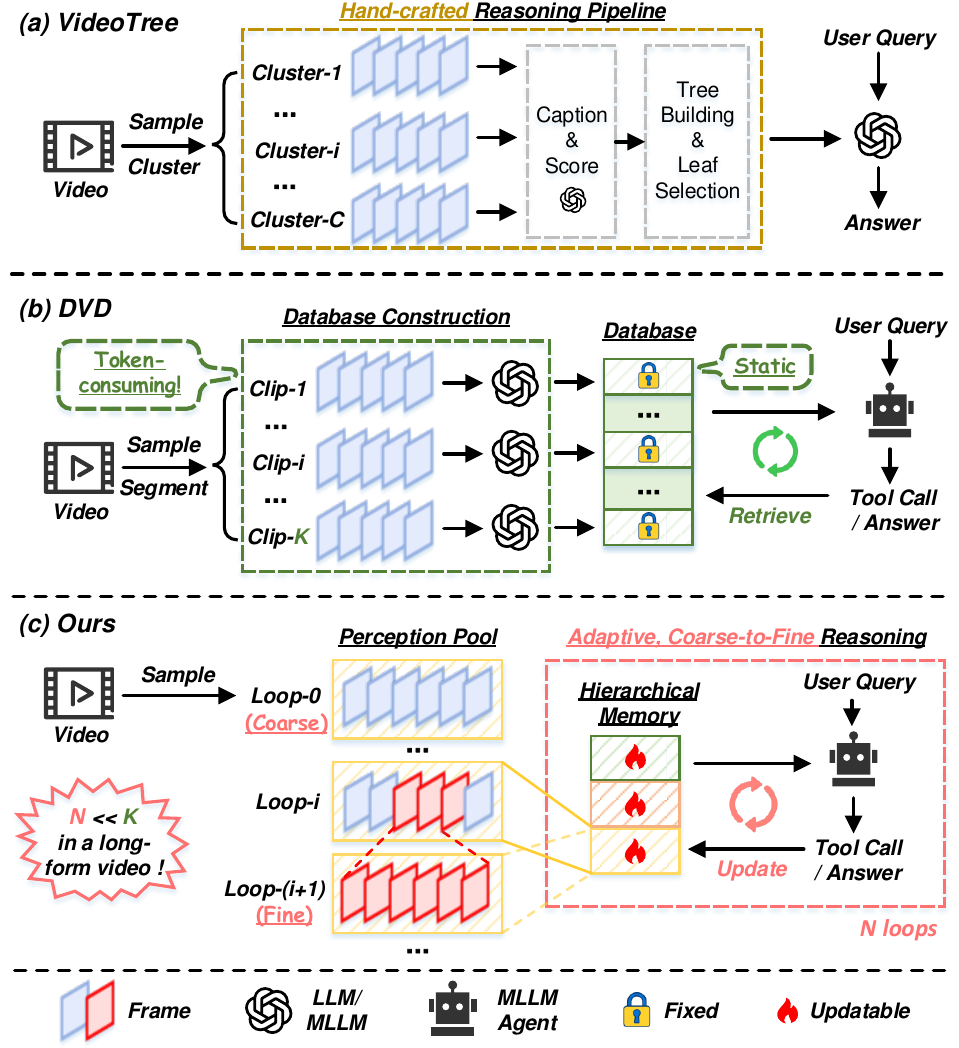}
    \vspace{-0.5em}
    \caption{Comparisons with LLM-based video understanding methods. Unlike the hand-crafted reasoning pipeline in VideoTree ~\cite{wang2025videotree} and the token-consuming static database construction in DVD \cite{zhang2025deep}, our VideoARM performs adaptive, coarse-to-fine video reasoning over a dynamically constructed hierarchical memory, enabling more efficient and flexible long-form video understanding within strict token constraints.}
    \label{fig:intro}
    \vspace{-1.2em}
\end{figure}

Many previous LLM (MLLM)-based methods~\cite{wang2024videoagent,wang2025videotree} rely on hand-crafted reasoning pipelines.
LLoVi~\cite{zhang2024simple} operates through a simple two-stage framework, dissecting long-range video question answering into short-term visual captioning and subsequently employing LLM for reasoning.
VideoTree~\cite{wang2025videotree} (Figure~\ref{fig:intro}~(a)) clusters video segments, generates and scores their captions to build a hierarchical tree structure, which is then fed into an LLM for query reasoning. 
Collectively, these manually gated designs limit autonomy and blunt the gains from stronger base LLM/MLLMs.

To tackle these challenges, DVD~\cite{zhang2025deep} constructs a clip-level database and employs a ReAct-like~\cite{yao2023react} agent to autonomously invoke tools for long-form video understanding, as shown in Figure~\ref{fig:intro}~(b). While effective, DVD suffers from two limitations: (1) for each video, DVD pre-segments the stream into 10-second clips and adopts MLLM to analyze each clip and construct a database, which is highly token-consuming and introduces substantial redundancy irrelevant to the query; (2) during agent execution, DVD embeds the query and retrieves from the pre-built database to localize key regions. This retrieval-centric paradigm fails to fully utilize the native visual reasoning capability of MLLMs, and the database cannot be updated on the fly to support the dynamics of observation and reflection during video understanding.

We introduce an \underline{\textbf{A}}gentic \underline{\textbf{R}}easoning-over-hierarchical-\underline{\textbf{M}}emory paradigm for long-form video understanding (VideoARM) to address these issues. VideoARM performs adaptive, coarse-to-fine agentic reasoning over a dynamically constructed hierarchical memory, discarding static, exhaustive preprocessing (Figure~\ref{fig:intro}~(c)).
To operationalize the agentic reasoning paradigm, VideoARM first provides two complementary toolsets that implement coarse-to-fine search and evidence refinement: (1) \textit{Temporal Scoping Tools}, which continuously or temporarily limit the agent’s focus to query-aligned segments, thereby significantly reducing its token consumption for processing irrelevant information. (2) \textit{Multimodal Understanding Tools}, which progressively localize, interpret, and abstract query-relevant evidence at either global or fine-grained levels. Combining these tools enables exploration across a broad temporal range, while also enabling rapid and low-overhead hypothesis testing and evidence gathering at the micro-event level.

To transform local observations into persistent, query-aware context, VideoARM complements the toolsets with a \underline{\textbf{h}}ierarchical \underline{\textbf{m}}ulti\underline{\textbf{m}}odal \underline{\textbf{m}}emory (HM$^3$) that is dynamically constructed and evolves throughout execution. 
HM$^3$ organizes memory into three tiers that, respectively, store current perceptual intervals, record intermediate tool outcomes in temporal sequence, and log the controller’s reasoning traces and objectives, transitioning from perception to cognition. 
Building on HM$^3$ and the toolsets above, the \textit{controller} adaptively schedules and invokes tools based on multimodal evidence stored in HM$^3$, fully leveraging multimodal planning and reasoning to achieve dynamic, context-aware decision-making throughout the video understanding process. Ultimately, VideoARM performs an adaptive and continuous loop of \textit{observing, thinking, acting, and memorizing}, thereby enabling effective, comprehensive long-form video understanding.

We conduct extensive experiments on prevalent long-form video understanding datasets, \textit{i.e.}, \textit{Video-MME}~\cite{fu2025video}, \textit{LongVideoBench}~\cite{wu2024longvideobench}, and \textit{EgoSchema}~\cite{mangalam2023egoschema}, across  different temporal regimes to validate the effectiveness of our VideoARM. Notably, compared with the previous state-of-the-art  method DVD~\cite{zhang2025deep}, our approach achieves better performance while significantly reducing token consumption for long-form videos.

\section{Related Work}
\label{sec:relatedwork}

\paragraph{Video MLLMs.}

Recent advances in multimodal large language models (MLLMs)~\cite{liu2023visual,ye2024mplug,shao2025growing,yu2025prophet,shao2025imp,openai2025o3,achiam2023gpt,openai2025gpt4.1,bai2025qwen2,damonlpsg2025videollama3,wang2025internvideo2} have significantly improved video understanding. Existing research primarily focuses on three aspects: 
\textbf{Context extension.} These methods~\cite{team2023gemini,wei2025visual,zhang2024long} expand context windows to handle long-video understanding, but they struggle with redundant frame information and high computational cost. \textbf{Token compression.} These methods~\cite{yang2025pvc,tao2025dycoke,shen2024longvu} dynamically compress redundant visual tokens while preserving relevant features. However, this compression can lead to the loss of fine-grained details, impacting tasks requiring precise visual reasoning. \textbf{Audio integration.} These methods~\cite{cappellazzo2025large,shu2025audio,xu2025qwen2} integrate dedicated audio encoders to enhance multimodal reasoning, but this increases model complexity and training costs. 
\vspace{-10pt}

\label{method}
\begin{figure*}
    \centering
    \includegraphics[width=0.98\linewidth]{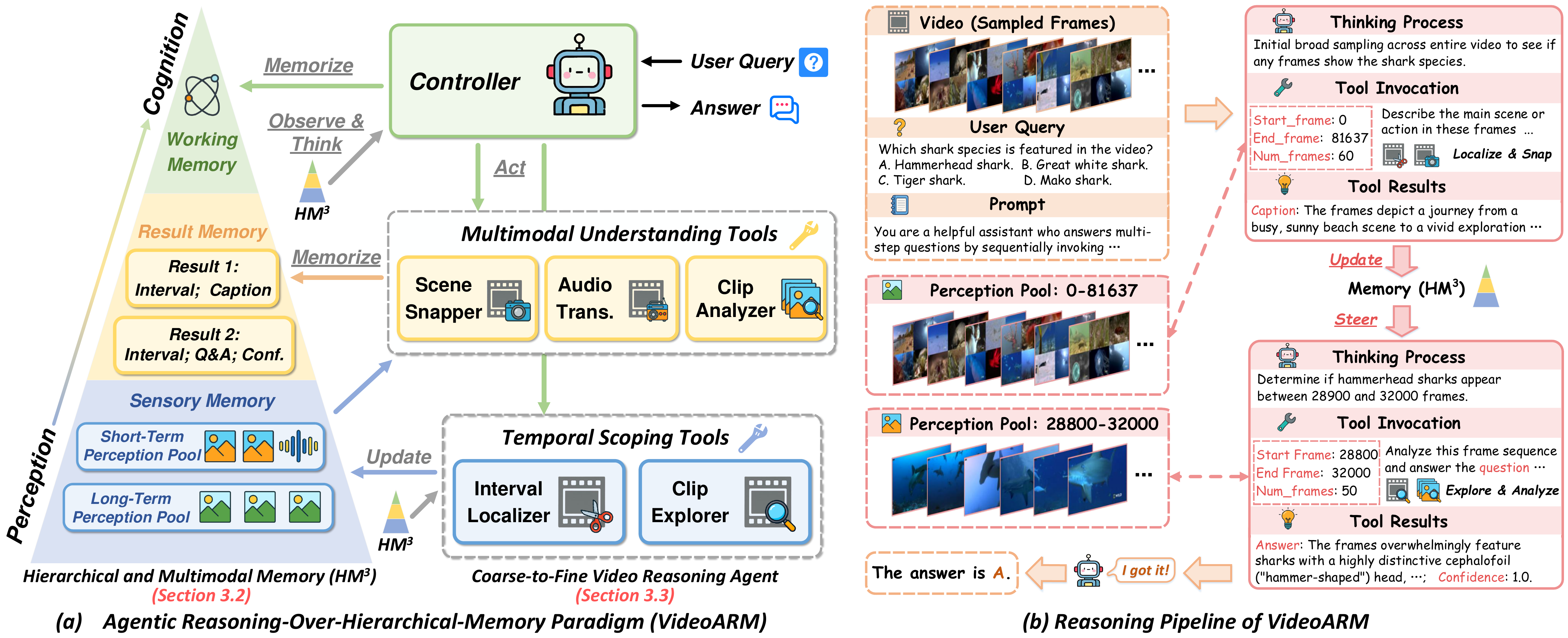}
    \vspace{-0.3em}

    \caption{Overview of the agentic reasoning-over-hierarchical-memory paradigm (VideoARM) and an example of its reasoning pipeline.
    }
    \label{fig:framework}
    \vspace{-0.6em}
\end{figure*}

\vspace{-10pt}
\paragraph{Video Agent.}
Agent-based methods~\cite{zuo2025videolucy, wang2024videoagent, fan2024videoagent, chen2025lvagent,li2024searchlvlms,zhao2024longagent} leverage the multimodal capabilities of LLM/MLLMs to integrate visual, audio, and text data, significantly enhancing the ability of frameworks to understand complex video content and reason across modalities for more accurate and robust interpretations.
Some approaches~\cite{he2024ma,wang2025videotree,yang2025vca} 
treat MLLMs as simple retrieval or reasoning components and rely on manually designed strategies to identify keyframes or maintain fixed memory structures. 
However, these methods heavily depend on predefined configurations, which constrain the autonomous reasoning and adaptability of LLM/MLLMs when processing long videos. 
Another category of methods~\cite{zhang2025deep,pang2025mr,jeong2025videorag,chen2025lvagent} 
develops an agentic search paradigm, where agents dynamically interact with clip-level databases or retrieved video segments. 
Deep Video Discovery (DVD)~\cite{zhang2025deep} introduces a ReAct-like~\cite{yao2023react} agent that autonomously invokes tools for clip-level reasoning. 
Nevertheless, these approaches remain limited by token-intensive pipelines and their reliance on prebuilt databases. 
Unlike these approaches, VideoARM replaces static, exhaustive preprocessing with adaptive, on-the-fly reasoning, while simultaneously dynamically storing contextual signals in memory. 
Moreover, compared to concurrent memory-based agents like VideoLucy~\cite{zuo2025videolucy} that rely on text summaries and fixed backtracking, VideoARM maintains a hierarchical buffer of raw multimodal evidence within an active “observe-think-act-memorize” loop.

\section{Method}

\subsection{Overview}
     VideoARM operates through an iterative  loop of \textit{observe–think–act–memorize}, where an agent decomposes user queries, determines the next focus of inspection, and coordinates multimodal tools. Simultaneously, a hierarchical multimodal memory continuously captures and organizes query-aware clues into a structured substrate. As illustrated in Figure \ref{fig:framework} (a), VideoARM consists of two core components: (i) a hierarchical and multimodal \textit{memory} HM$^3$; (ii) a coarse-to-fine video reasoning agent managed by a  \textit{controller} that adaptively schedules \textit{tool} invocation and guides decision-making over long temporal horizons.

\subsection{Hierarchical and Multimodal Memory}
    The \textbf{H}ierarchical and \textbf{M}ulti\textbf{m}odal \textbf{M}emory (HM$^3$) serves as the contextual knowledge base of VideoARM, and is continuously updated during agent execution. 
    It is organized into three tiers based on semantic hierarchy: 

    \noindent\textbf{\memoryA\xspace Sensory Memory.}
        The Sensory Memory consists of a long-term perception pool $P_l$ and a short-term perception pool $P_s$ for storing perceptual information, such as key frames $\mathcal{F}$. 
        The long-term perception pool $P_l$ maintains a snapshot of the base frame interval currently processed by the pipeline, consisting of frames within that temporal segment. It acts as a lightweight, volatile buffer that captures the agent’s most recent temporal focus.
        In contrast, the short-term perception pool $P_s$ temporarily holds fine-grained clips sampled during local exploration and analysis. It serves as a transient workspace for tools, enabling rapid hypothesis testing and evidence accumulation at the micro-event level. Once the analysis results are written into \emph{Result Memory} (detailed below), the short-term perception pool is cleared to free resources for the next reasoning cycle.
        
    \noindent\textbf{\memoryB\xspace Result Memory.}
        The Result Memory serves as the dynamic workspace that records the intermediate results generated throughout the agent’s operation. It stores the iteration index, analyzed frame intervals, and the corresponding tool outputs. By maintaining these temporally ordered entries, the Result Memory captures a fine-grained history of the agent’s perception cues. This allows the controller to reflect on past reasoning steps, avoid redundant actions, and adaptively adjust its subsequent strategy based on accumulated evidence.  

    \noindent\textbf{\memoryC\xspace Working Memory.}
        The Working Memory primarily records the controller’s reasoning traces and intended objectives prior to each tool invocation. By externalizing these intermediate traces from the MLLM’s context into Working Memory, we can clear tool-related context after every call. This design reduces redundancy in the model’s context, mitigates context overflow, and sharpens the focus and efficiency of subsequent reasoning. In addition, the persisted traces provide an auditable history that supports error analysis and strategy adaptation.

    Overall, the raw perceptual inputs stored in \emph{Sensory Memory} \memoryA\xspace are progressively transformed into fine-grained semantic evidence in \emph{Result Memory} \memoryB\xspace through the use of tools, while deeper semantic information, specifically the \textit{controller}’s reasoning traces and intentions, is recorded in \emph{Working Memory} \memoryC\xspace. Together, these components constitute a dynamically evolving and hierarchical memory.

\subsection{Coarse-to-Fine Video Reasoning Agent}
    Prior works commonly follow a bottom-up hierarchical pipeline in which fixed, fine-grained memories are preconstructed and subsequently queried by an MLLM for retrieval and reasoning \cite{fan2024videoagent, zhang2025deep}. In contrast, by leveraging HM$^3$, our agent adopts a query-guided, coarse-to-fine reasoning pipeline that gradually decomposes global semantics into local evidence. Our agent consists of two main components: (i) \textit{multimodal toolsets} used to determine the range of video segments being processed and to gather query-relevant evidence from different perspectives. (ii) a \textit{controller} adaptively reasons, schedules, and invokes tools based on the multimodal evidence stored in HM$^3$.
    
\subsubsection{Multimodal Toolsets}
    To support this coarse-to-fine reasoning pipeline, we design two complementary toolsets: \textit{Temporal Scoping Tools} \toolA\xspace, which localizes and probes query-relevant temporal regions, and \emph{Multimodal Understanding Tools} \toolB\xspace, which summarizes, transcribes, and analyzes local content to provide fine-grained evidence. 
    Given a video $V$ and its corresponding audio, these tools enable progressive refinement and evidence aggregation from global to local levels. We specify each tool in these two toolsets as follows, with the icon preceding each tool indicating its corresponding toolset.

    \noindent\textbf{\toolA\xspace Interval Localizer.} The Interval Localizer identifies the frame intervals $T_{long} = [t_i, t_j]$ in a video that are most relevant to the user query by leveraging contextual signals stored in HM$^3$.
    In addition, it determines the sampling frame number $N_1$ (\textit{e.g.}, 150) for each video interval, thereby enabling adaptive granularity control over subsequent visual analysis. 
    Next, it uniformly samples $N_1$ frames from the video within $T_{long}$. To accommodate the visual context limits of the MLLM, the sampled frames are composited into compact 3×2 image grids, with each tile overlaid by its frame index.
    Finally, it uses these image grids to update the long-term perception pool within \emph{Sensory Memory}  \memoryA\xspace. 
    Through continuous evidence integration over the stored information, the tool adaptively constrains the agent’s working scope to these query-aligned segments and their surrounding context, thereby significantly narrowing the attention range and reducing token consumption during the reasoning process.

    \noindent\textbf{\toolA\xspace Clip Explorer.} The Clip Explorer serves a similar role to the \emph{Interval Localizer}, but does not alter the agent’s focused segments \textit{(i.e.}, long-term perception pool). Instead, it performs tentative, fine-grained probing within local frame intervals $T_{local}$ around the focused segments. Given the short duration of the frame intervals it focuses on, we directly store the frames corresponding to these intervals in the short-term perception pool within \emph{Sensory Memory} \memoryA\xspace, without using a tiling method, and employ a fixed sampling frame number $N_2$. We also store the audio segment $\mathcal{A}$ within intervals in the short-term perception pool. Overall, this tool maintains the agent’s global focus while enabling rapid, low-overhead hypothesis testing and evidence gathering at the micro-event level.

    \noindent\textbf{\toolB\xspace Scene Snapper.} Scene Snapper specializes in describing long temporal segments.  This tool analyzes and summarizes the frames $\mathcal{F}$ stored in the long-term perception pool $P_l$, generating a concise caption $V_C$ that describes the main scene. This process provides a compact yet informative abstraction of long-term visual content, facilitating high-level semantic reasoning and efficient memory storage in HM$^3$.
    \begin{equation}
        V_C = \text{SceneSnapper}(\mathcal{F}), \quad \mathcal{F} \in P_l.
    \label{eq:scene_snapper}
    \end{equation}
    \noindent\textbf{\toolB\xspace Audio Transcriber.} When local visual cues are insufficient or ambiguous, this tool invokes whisper-1 \cite{radford2023robust} to extract the audio segment $\mathcal{A}$ from the short-term perception pool $P_s$ and generate a transcript $A_C$, thereby providing semantic evidence that complements the visual signals.
    \begin{equation}
        A_C = \text{AudioTrans}(\mathcal{A}), \quad \mathcal{A} \in P_s.
    \label{eq:audio_transcripter}
    \end{equation}
    \noindent\textbf{\toolB\xspace Clip Analyzer.}
        Clip Analyzer is focused on capturing fine-grained local semantic details. Given the frames $\mathcal{F}$ stored in short-term perception pool $P_s$ and a sub-question $Q_{sub}$, this tool jointly examines these frames, synthesizing cross-clip observations to infer detailed spatial and semantic information. It then produces an answer $A_{sub}$ to the sub-question along with a confidence score $S_{sub}$ that reflects the reliability of its reasoning. This tool enables precise semantic discrimination and temporal understanding within local segments, providing essential fine-grained evidence and validation support for long-form video reasoning.
    \begin{equation}
        A_{sub}, \ S_{sub} = \text{ClipAnalyzer}(\mathcal{F}, \ Q_{sub}), \quad \mathcal{F} \in P_s.
    \label{eq:clip_analyzer}
    \end{equation}
    After the execution of the \emph{Multimodal Understanding Tools}, we update the results along with their corresponding temporal intervals to the \emph{Result Memory} \memoryB\xspace, serving as localized semantic evidence for subsequent reasoning. The short-term perception pool is cleared afterward.

\subsubsection{Controller}
    To maximize the intrinsic reasoning and planning capacity of modern MLLMs, we intentionally avoid rigid workflows and pre-defined tool-usage rules. The controller operates in a streamlined observe–think–act–memorize loop, akin to ReAct-style interaction \cite{yao2023react}, but grounded by HM$^3$ as a persistent and hierarchical memory. For a given user query, the controller reasons over the current observation state (summaries and cues stored in HM$^3$), selects an appropriate tool and its parameters, executes it, and writes the newly acquired evidence back to memory. This loop continually updates the context, sharpens internal hypotheses, and progressively narrows the search space.
    
    \noindent\textbf{Action space and state. } Let the initial query be $Q$, the tool set $TS$, and the terminal action \textsc{Answer}. Considering the correspondence between the \emph{Temporal Scoping Tools} \toolA\ and the \emph{Multimodal Understanding Tools} \toolB, we instantiate the tool set $TS$ using only the latter. The controller maintains an interaction history $H_t$ and reads from HM$^3$’s three tiers. At each iteration $t$, it forms a reasoning trace $R_t$, chooses an action $A_t$ in $TS \ \cup$ \{\textsc{Answer}\}  with parameters $P_t$ (\textit{e.g.}, sub-question $Q_{sub}$) and its corresponding \emph{Temporal Scoping Tool} \toolA\xspace (if not \textsc{Answer}), obtains observations $O_t$ (\textit{e.g.}, captions $V_C$), and writes them back to the HM$^3$.

    \noindent\textbf{Iterative reasoning loop. } Concretely, the controller follows an iterative reasoning loop. It first \textbf{\textit{observes}} the current global-to-local contextual information stored in HM$^3$, and \textbf{\textit{thinks}} by generating a compact plan $R_t$ that hypothesizes what evidence is missing and which modality would be most informative next. Based on this reasoning, it \textbf{\textit{acts}} by invoking one tool with parameters $P_t$. After obtaining the tool’s output, it \textbf{\textit{memorizes}} by appending $O_t$ to the \emph{Result Memory} \memoryB\xspace. In addition, the controller’s reasoning trace or intent for invoking this tool is recorded into the \emph{Working Memory} \memoryC\xspace. Finally, it refines its internal hypothesis, and if uncertainty remains high, the controller repeats the cycle from the observation step. Specially, the process stops when the controller either selects \textsc{Answer} in the \textit{act} stage or reaches the step budget $N$, and generates the final answer. An example of the reasoning pipeline of VideoARM is shown in Figure \ref{fig:framework} (b).

    \noindent\textbf{Design principles.} By centering the MLLM’s reasoning in the loop and coupling it with HM$^3$’s dynamically constructed hierarchical memory, the controller acquires an autonomous, evidence-guided, and adaptive action policy. It balances exploitation—leveraging \textit{Audio Transcriber} and \textit{Clip Analyzer} for targeted, high-precision semantic analysis and verification within localized intervals—and exploration—using \textit{Scene Snapper} to gather coarse-to-fine evidence over long temporal spans, refresh HM$^3$ summaries, and surface promising regions for subsequent probing. This coordination tightens local inference while preserving broad temporal coverage and token efficiency. Empirically, the query-oriented control mechanism substantially reduces token consumption while maintaining, and often improving, temporal coherence and final-answer accuracy on long-form video reasoning tasks.

\section{Experiment}

\begin{table*}[t]
    \centering
    \begin{minipage}[t]{0.70\textwidth}
    \vspace{0pt}
    \renewcommand\arraystretch{1.0}
    \resizebox{1.00\linewidth}{!}{
    \begin{threeparttable}
        \begin{tabular}{l ccccccc}
            \toprule
            \multirow{2}{*}{\textbf{Methods}} &  \multicolumn{4}{c}{\textbf{Video-MME (w/o sub)}}    &  \multicolumn{2}{c}{\textbf{LongVideoBench}}   &  \multirow{2}{*}{\textbf{EgoSchema}} \\
            \cmidrule(lr){2-5} \cmidrule(lr){6-7}
            ~   & \textit{Short} & \textit{Medium} & \textit{Long}  & \textit{Overall} & \textit{Long} & \textit{Overall}  &  \\
            \midrule
            \multicolumn{4}{l}{\textcolor{gray}{\textit{Commercial VLMs}}} \vspace{0.2mm} \\
            Gemini-1.5-Pro~\cite{team2023gemini}  & 81.7 & 74.3 & 67.4 & 75.0 & 58.6 & 64.0 & 71.1\\
            Gemini-2.0-Flash~\cite{team2023gemini}    & - & - & 63.0 &  & 45.7 & - & 71.2 \\
            GPT-4o~\cite{achiam2023gpt}  & 80.0 & 70.3 & 65.3 & 71.9 & 60.9 & 66.7 & 72.2 \\
            OpenAI o3~\cite{openai2025o3} & - & - & 63.2 & - & 60.6 & 67.5 & 63.2\\
            \midrule
            \multicolumn{4}{l}{\textcolor{gray}{\textit{Open-Source VLMs}}} \vspace{0.2mm} \\
            mPLUG-Owl3~\cite{ye2024mplug} &  70.0 & 57.7 & 50.1 & 59.3 & - & 59.8 &  - \\
            InternVL2.5-72B~\cite{wang2025internvideo2}   & 82.8 & 70.9 & 62.6 & 72.1 & - & 63.6 & - \\
            Qwen2.5-VL-72B~\cite{bai2025qwen2}    & - & - & - & 73.3 & - & 60.7 & 	76.2 \\
            AdaReTaKe~\cite{wang2025adaretake}  & 80.6 & 74.9 & 65.0 & 73.5 & - & 67.0 &  - \\
            \midrule
            \multicolumn{4}{l}{\textcolor{gray}{\textit{Other Training-Free Approaches}}} \vspace{0.2mm} \\
            VideoAgent~\cite{wang2024videoagent}  & - & - & - & - & - & - &  60.2 \\
            VideoAgent~\cite{fan2024videoagent}  & - & - & - & - & - &  & 63.2 \\
            VideoTree~\cite{wang2025videotree}  & - & - & 54.2 & - & - & - &  67.0 \\
            VCA~\cite{yang2025vca}   & - & - & - & - & - & - & 73.6 \\
            Logic-in-Frames~\cite{guo2025logic}  & 71.9 & 61.9 & 55.2 & 63.0 & 54.2 & - & - \\
            MR. Video~\cite{pang2025mr}   & - & - & 61.8 & - & 61.6 & - & 73.0 \\
            VideoLucy~\cite{zuo2025videolucy}   & 78.6 & 72.1 & 66.8 & 72.5 & - & - & - \\
            DVD~\cite{zhang2025deep}   & - & - & 67.3 & - & 68.6 & 71.6 & 
            \underline{76.6} \\
            \midrule
            \multicolumn{4}{l}{\textcolor{gray}{\textit{VideoARM (ours)}}} \vspace{0.2mm} \\
            \textbf{VideoARM} (o3+GPT-4.1)  & \textbf{86.4} & \underline{78.4} & \underline{75.3} & \underline{80.1} & \underline{69.2} & \underline{73.7} & \textbf{78.2} \\
            \textbf{VideoARM} (o3+GPT-4o)  & \underline{85.8} & \textbf{81.3} & \textbf{81.2} & \textbf{82.8} & \textbf{76.4} & \textbf{78.0} & 76.2 \\
            \bottomrule
        \end{tabular}
    \end{threeparttable}
    }
\end{minipage}
\hspace{0.2em}
\begin{minipage}[t]{0.28\textwidth}
\vspace{0pt}
\centering

\begin{subtable}[t]{\linewidth}
\centering
\resizebox{\linewidth}{!}{%
\begin{tabular}{lc}
    \toprule
    \textbf{Method} & \textbf{MLVU} \\
    \midrule
    \textcolor{gray}{\textit{Commercial VLMs}} & \\
    GPT-4V~\cite{achiam2023gpt} & 49.2 \\
    GPT-4o~\cite{achiam2023gpt} & 64.6 \\       
    \midrule
    \textcolor{gray}{\textit{Open-Sourced MLLMs}} &  \\ 
    LLaVA-OV-72B~\cite{li2024llavaov} & 66.4 \\  
    VideoChat-Flash-7B~\cite{li2024videochat} & 74.7 \\  
    \midrule
   \textcolor{gray}{\textit{Training-Free Approaches}} \\
    VideoTree~\cite{wang2025videotree} & 60.4\\
    VideoLucy~\cite{zuo2025videolucy} & \underline{76.1} \\  
    \textbf{VideoARM} (o3+GPT-4.1) & \textbf{81.2} \\  
    \bottomrule
\end{tabular}}
\end{subtable}

\vspace{1.0mm}

\begin{subtable}[t]{\linewidth}
\centering
\resizebox{\linewidth}{!}{%
\begin{tabular}{lc}
    \toprule
    \textbf{Method} & \textbf{LVBench} \\
    \midrule
    \textcolor{gray}{\textit{Commercial VLMs}} & \\
    Gemini-1.5-Pro~\cite{team2023gemini} & 33.1 \\
    GPT-4o~\cite{achiam2023gpt} & 48.9 \\       
    \midrule
    \textcolor{gray}{\textit{Open-Sourced MLLMs}} &  \\ 
    Qwen2-VL-72B~\cite{wang2024qwen2} & 41.3 \\  
    AdaReTaKe~\cite{wang2025adaretake} & 53.3 \\  
    \midrule
   \textcolor{gray}{\textit{Training-Free Approaches}} \\
    VideoLucy~\cite{zuo2025videolucy} & 58.8 \\  
    DVD~\cite{zhang2025deep} & \underline{76.0} \\
    \textbf{VideoARM} (o3+GPT-4.1) & \textbf{79.7} \\  
    \bottomrule
\end{tabular}}
\end{subtable}
\end{minipage}

\vspace{-1mm}
\caption{Comparison with \textit{commercial VLMs}, \textit{open-source VLMs}, and other \textit{training-free approaches}  on prevalent long video understanding benchmarks. Results are reported as accuracy (\%) on different temporal regimes (short, medium, long) and overall.}
\label{tab:long_video_benchmarks}

\vspace{-4.5mm}
\end{table*}

\subsection{Benchmarks}
We evaluate our method on several prevalent long-form video understanding benchmarks, including Video-MME~\cite{fu2025video}, LongVideoBench~\cite{wu2024longvideobench}, and EgoSchema~\cite{mangalam2023egoschema}. These datasets cover a wide range of temporal scales, scene complexities, and multimodal challenges, providing a comprehensive evaluation of long-form video reasoning.

\noindent\textbf{Video-MME}~\cite{fu2025video} includes 900 videos, totaling 254 hours and 2,700 multimodal question–answer pairs. The video durations range from 11 seconds to 1 hour, with an average duration of 1,024 seconds. 
This benchmark evaluates both short- and long-term temporal reasoning and integrates visual, audio, and text modalities. We report all results under the ``w/o subtitles" (denoted as ``w/o sub") setting, where subtitles of videos are unavailable.

\noindent\textbf{LongVideoBench}~\cite{wu2024longvideobench} features 6,678 questions from 3,763 videos, ranging in duration from a few seconds to an hour. We emphasize the longest subset with durations in $(900s, 3600s]$ (\textit{i.e.}, \textit{long} subset), comprising 564 questions from 188 videos. In our experiments, we evaluate the validation split consisting of 1,337 question–answer pairs, which provides a balanced coverage across various durations and scene complexities. 

\noindent\textbf{EgoSchema}~\cite{mangalam2023egoschema} consists of over 5,031 human-curated question–answer pairs derived from the Ego4D dataset. Each question corresponds to a three-minute egocentric video clip, requiring the model to reason over extended temporal and contextual information. To ensure efficient yet representative evaluation, we utilize a subset of 500 question–answer pairs for testing.

\subsection{Implementation Details}
VideoARM follows an agentic, query-guided reasoning paradigm.
We conduct all experiments by accessing multimodal large language model (MLLM) via API. We use OpenAI o3 to implement the \textit{Controller} and the \textit{Temporal Scoping Tools}, and whisper-1 model \cite{radford2023robust} to implement the \textit{Audio Transcriber}. For fair comparison, we adopt two configurations (GPT-4.1 or GPT-4o) to implement the \textit{Scene Snapper} and \textit{Clip Analyzer}. Specially, considering the cost factor, we use GPT-4.1 for these two tools in our ablation experiments unless otherwise specified.
For \textit{Interval Localizer}, the sampling frame number $N_1$ ranges from 30 to 150. Each sampled frame is resized with its short edge set to 256, and a tiling strategy is applied to preserve visual details. 
For \textit{Clip Explorer}, the short edge is set to 512 to capture fine-grained spatial details, and the stored audio segments are constrained to 25 MB, due to the limitations of the whisper-1 model.
We set the step budget $N$ to 10 to balance performance and efficiency.

\subsection{Main Results}
\begin{figure}
    \centering
    \includegraphics[width=0.99\linewidth]{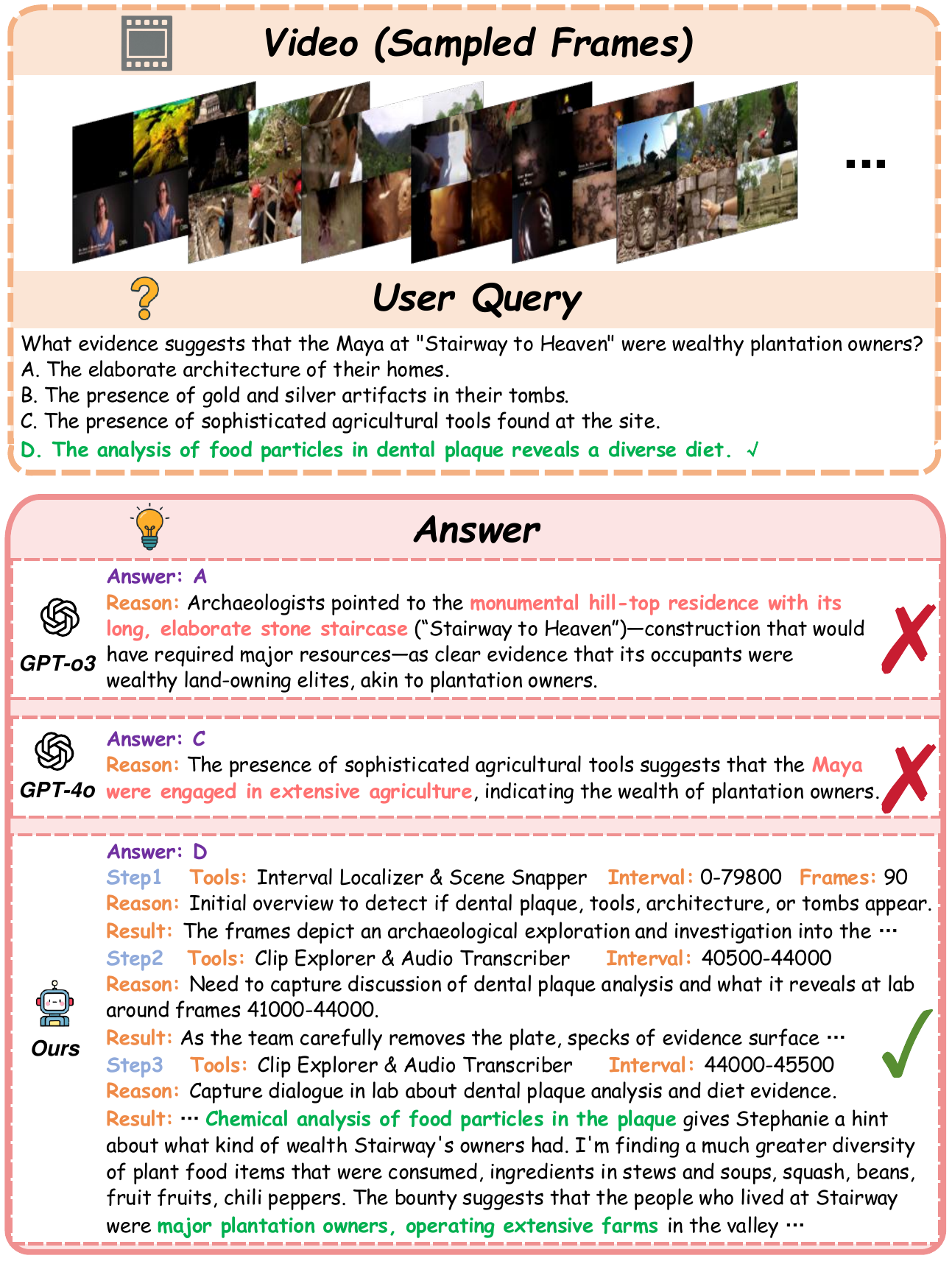}  
    \vspace{-0.6em}
    \caption{Visualization of the responses and underlying reasoning processes generated by different methods to a specific question. 
    }
    \label{fig:cmp}
    \vspace{-0.8em}
\end{figure}

Table~\ref{tab:long_video_benchmarks} presents a comprehensive comparison across several prevalent long-video benchmarks.
On Video-MME, VideoARM consistently outperforms prior methods, including commercial VLMs, open-source VLMs, and other training-free approaches,  across all evaluation metrics.
VideoARM also achieves new state-of-the-art performance across the LongVideoBench, EgoSchema, MLVU, and LVBench benchmarks.
Moreover, our method yields substantial gains over its base models (GPT-4o and OpenAI o3) on the first three benchmarks, demonstrating the effectiveness of our VideoARM in enhancing long-term multimodal understanding.
Taking the example shown in Figure \ref{fig:cmp}, GPT-4o and OpenAI o3 fail to answer the question, while our method progressively identifies key temporal intervals and retrieves relevant information through the autonomous invocation of tools, ultimately providing the correct answer.

For a fair comparison with DVD~\cite{zhang2025deep}, we adopt the same backbone configuration (OpenAI o3+GPT-4.1) and still outperform DVD across all metrics. This demonstrates that our method can achieve better performance while maintaining low token consumption (as discussed in Section \ref{sec:token}), further validating the effectiveness of our approach.

These results highlight several key advantages of our framework. First, the controller adaptively focuses on query-relevant temporal segments in a coarse-to-fine manner, while the hierarchical multimodal memory enables efficient aggregation and reasoning over visual, audio, and textual cues. This leads to a more principled and effective use of video modalities compared to prior approaches. Second, by leveraging a training-free, tool-augmented pipeline, VideoARM achieves state-of-the-art performance without any additional model fine-tuning, thereby markedly lowering both computational and memory costs.
In summary, VideoARM demonstrates that effective video understanding can be achieved in a computation- and modality-efficient manner, offering a practical and generalizable solution for both long-form and short-form video comprehension tasks.

\begin{table}[t]
    \vspace{1mm}
    \centering
    \renewcommand\arraystretch{1.3}
    \setlength{\tabcolsep}{2.5pt}
    \rowcolors{6}{gray!20}{gray!20}
    \resizebox{1\linewidth}{!}{
    \begin{threeparttable}
        \begin{tabular}{c|ccc|c}
        \toprule
        \multirow{2}{*}{\textbf{Controller}} & \multicolumn{3}{c|}{\textbf{Tools}} & \textbf{Video-MME} \\
        ~ & \textit{Temporal Scop.}  & \textit{MM Und.*} & \textit{Audio Trans.} &  \textbf{(long)} \\\midrule
        Qwen3-VL     & Qwen3-VL        & GPT-4.1   & \multirow{1}{*}{whisper-1}  & 54.9 \\
        GPT-4o       & GPT-4o        & GPT-4.1   &  whisper-1  & 40.5 \\ 
        GPT-5        & GPT-5        & GPT-4.1   &  whisper-1  & 75.5 \\ 
        OpenAI o3           & OpenAI o3        & GPT-4.1   & whisper-1 & 76.5 \\  
        OpenAI o3           & OpenAI o3         & GPT-4o    &   whisper-1 & \textbf{80.0}  \\ 
        \bottomrule
        \end{tabular}
    \end{threeparttable}
    }
    \vspace{-0.6em}
    \caption{Ablation on model choices for the controller and toolsets. ``\textit{Temporal Scop.}" denotes \textit{Temporal Scoping Tools}, ``\textit{MM Und.*}" denotes \textit{Multimodal Understanding tools} (excluding \textit{Audio Transcriber}), and ``\textit{Audio Trans.}" denotes the \textit{Audio Transcriber}. }
    \vspace{-1.2em}
    \label{tab:ablation-vlms}
\end{table}

\subsection{Ablation Study}
For all the ablation studies, we selected 200 question-answer pairs from each used dataset as the evaluation subset, based on a stratified sampling strategy that preserves the original distribution of video types and task categories.

\subsubsection{Model Choices}
We systematically vary the controller and the toolsets to assess their impact on video multimodal reasoning, as shown in Table \ref{tab:ablation-vlms}.
Considering that both the controller and the \textit{Temporal Scoping Tools} primarily perform decision-making and reasoning, we instantiate them with the same model; and because whisper-1 is the only officially available audio transcription tool from OpenAI, we fix whisper-1 as the \textit{Audio Transcriber} tool across all experiments. 
Using OpenAI o3 as the controller and GPT-4o for \textit{Multimodal Understanding Tools} (excluding \textit{Audio Transcriber}) achieves the best score, 80.0\%. Holding the controller fixed to OpenAI o3, replacing GPT-4.1 with GPT-4o in \textit{Multimodal Understanding Tools} yields a +3.5\% gain, indicating that GPT-4o possesses enhanced multimodal capabilities for captioning and semantic analysis. With \textit{Multimodal Understanding Tools} fixed to gpt-4.1, switching the controller from GPT-5 to OpenAI o3 offers a smaller improvement (+1.0\%), placing both in the top tier. In contrast, configurations with Qwen3-VL and GPT-4o as controllers score 54.9\% and 40.5\%, respectively, substantially below the OpenAI o3/GPT-5 group, suggesting limitations in their reasoning and decision-making capabilities, as GPT-5 and OpenAI o3 exhibit stronger temporal reasoning and hierarchical planning abilities, while GPT-4o and Qwen3-VL remain constrained in complex multi-step inference.

\begin{table}[t]
    \centering
    \renewcommand\arraystretch{1.15}
    \resizebox{\linewidth}{!}{
        \begin{threeparttable}
            \begin{tabular}{ccc|c}
            \toprule
            \multicolumn{3}{c|}{\textbf{Multimodal Understanding Tools}} & \textbf{Video-MME} \\ 
            \textit{Scene Snapper}   & \textit{Audio Transcriber}   & \textit{Clip Analyzer} & \textbf{(long)} \\ \midrule 
                        & \checkmark    & \checkmark    &  69.0 \\
            \checkmark  &               & \checkmark    &  70.5 \\
            \checkmark  & \checkmark    &               &  75.5 \\
            \rowcolor{gray!20}
            \textbf{\checkmark}  & \checkmark    & \checkmark    &  \textbf{76.5} \\
            \bottomrule
            \end{tabular}
        \end{threeparttable}
    }
    \vspace{-0.6em}
    \caption{Ablation on the \textit{Multimodal Understanding Tools}.}
    \label{tab:ablation-tools}
    \vspace{-1.2em}
\end{table}

\begin{table}[t]
\centering
\vspace{3mm}
\renewcommand\arraystretch{1.15}
\resizebox{\linewidth}{!}{
\begin{threeparttable}
\begin{tabular}{ccccc|c}
\toprule 
\multicolumn{4}{c}{\textbf{Memory Type} }  & \textbf{Controller} & \textbf{Video-MME} \\
\textit{Short-term} & \textit{Long-term} & \textit{Result} & \textit{Working} & \textit{Context} & \textbf{(long)}\\ 
\midrule
 & \checkmark & \checkmark & \checkmark &  & 72.5 \\  
\checkmark &  & \checkmark & \checkmark &  & 67.0 \\  
\checkmark & \checkmark &  & \checkmark &  & - \\  
\checkmark & \checkmark & \checkmark &  &  & 75.5 \\  
 &  &  &  & \checkmark &  74.5\\  
\rowcolor{gray!20}
\checkmark & \checkmark & \checkmark & \checkmark &  & \textbf{76.5} \\ 
\bottomrule
\end{tabular}
\end{threeparttable}
}
\vspace{-0.6mm}
\caption{Ablation on the \textit{Hierarchical Multimodal Memory}, where the \textit{Short-term} and \textit{Long-term} denotes the \textit{short-term} and \textit{long-term perception pool} in the \textit{Sensory Memory}, respectively. \textit{Context} denotes that the controller context is not refreshed.}
\label{tab:ablation-memory}
\vspace{-0.8em}
\end{table}

\subsubsection{Multimodal Understanding Tools}
We evaluate the contribution of three tools, \textit{i.e.}, Scene Snapper (SS), Audio Transcriber (AT), and Clip Analyzer (CA), by toggling each component, as shown in Table \ref{tab:ablation-tools}. Adding Scene Snapper (SS) on top of {AT, CA} improves the score from 69.0\% to 76.5\%, showing that long-range global summarization effectively guides downstream reasoning. Adding Audio Transcriber (AT) on top of {SS, CA} yields a 6.5\% gain, indicating that audio grounding resolves visually ambiguous or off-screen events. Adding Clip Analyzer (CA) on top of {SS, AT} improves the score from 75.5\% to 76.5\%, confirming that fine-grained local analysis supplies complementary detail-level evidence. The full configuration (SS+AT+CA) yields the best result (76.5\%), demonstrating the synergy among global summarization, audio cues, and local semantic verification.

\subsubsection{Hierarchical Multimodal Memory}
We assess the impact of different memory components within \textit{Hierarchical and Multimodal Memory}, as shown in Table \ref{tab:ablation-memory}. Excluding the \textit{short-term perception pool} causes a 4.0\% performance drop, highlighting the importance of quickly testing hypotheses and gathering fine-grained evidence. Removing the \textit{long-term perception tool} leads to a 9.5\% drop, as the agent loses its ability to narrow its attention, resulting in difficulty accurately locating relevant information. Without \textit{Result Memory}, the agent struggles to determine its next steps, causing repeated cycles and invalid results. Removing \textit{Working Memory} results in a 1.0\% drop, showing that previous reasoning traces contribute meaningfully to the reasoning process. Additionally, relying solely on the context within the controller, without HM$^3$, causes a 2.0\% performance drop, emphasizing HM$^3$’s role in capturing key information and relieving the MLLM’s contextual burden, thus improving overall reasoning ability.

\begin{table}[t]
\begin{minipage}[t]{0.98\linewidth}
    \renewcommand{\arraystretch}{1.1}
    \centering
    \resizebox{\textwidth}{!}{
    \begin{tabular}{c|ccc}
    \toprule
        \begin{tabular}[c]{@{}c@{}}\textbf{$N$} \end{tabular} 
        & \textbf{Video-MME(short)} & \textbf{Video-MME(long)} & \textbf{LongVideoBench} \\
        \midrule
        3 & \textbf{87.5} & 72.0 & 67.5 \\
        5 & 87.0 & 74.5 & 67.5 \\
        7 & 85.5 & 75.5 & 69.5 \\
        \rowcolor{gray!20}
        \textbf{10} & 84.0 & \textbf{76.5} & \textbf{70.5} \\
        \bottomrule
        \end{tabular}
    }
    \vspace{-0.6em}
    \caption{
    Ablation study on the step budget $N$.
    }
    \label{tab:threshold}
\end{minipage}
\vspace{-0.5em}
\end{table}

\begin{table}[t]
\begin{minipage}[t]{0.98\linewidth}
    \renewcommand{\arraystretch}{1.3}
    \centering
    \resizebox{\linewidth}{!}{
        \begin{tabular}{lcc}
        \toprule
        \textbf{Sampling Strategy} & \textbf{Video-MME(long)} & \textbf{LongVideoBench} \\
        \midrule
        Fixed $N_1=30$ &  73.5  &  68.0  \\
        Fixed $N_1=60$ &  74.0  &  \textbf{70.5} \\
        \rowcolor{gray!20}
        \textbf{Adaptive} $N_1$ (Avg.: 49.8) &  \textbf{76.5}  &  \textbf{70.5}  \\
        \bottomrule
        \end{tabular}
    }
    \vspace{-0.6em}
    \caption{Ablation study on the frame sampling strategy of the \textit{Interval Localizer} tool.}
    \vspace{-0.8em}
    \label{tab:ablation_sampling_N1}
\end{minipage}
\end{table}

\subsubsection{Step Budget}
We conduct an ablation study on the step budget $N$ to analyze its impact on performance. As shown in Table \ref{tab:threshold}, increasing $N$ from 3 to 10 leads to consistent improvements on long-video datasets, indicating that a larger reasoning budget allows the agent to perform deeper exploration and evidence refinement. However, the gains on short videos are marginal, suggesting that excessive reasoning steps may introduce redundancy when the temporal scope is limited. Overall, setting $N = 10$ achieves the best balance between performance and efficiency.

\subsubsection{Frame Sampling Strategy}
We compare fixed sampling strategies with an adaptive, controller-driven strategy for the \textit{Interval Localizer} tool across two long-video benchmarks, as shown in Table \ref{tab:ablation_sampling_N1}. Using a fixed strategy with $N_1=60$ outperforms $N_1=30$ on both datasets, indicating that denser coverage generally benefits long-horizon reasoning. The adaptive strategy achieves the best result on both datasets, despite using fewer frames on average (${N_1}=49.8$). These results suggest that adapting the sampling strategy to interval complexity helps capture salient events without over-sampling easy segments, improving both effectiveness and efficiency.

\begin{table}[t]
\begin{minipage}[t]{1.00\linewidth}
    \renewcommand{\arraystretch}{1.3}
    \centering
    \resizebox{\linewidth}{!}{
        \begin{tabular}{l|cc}
        \toprule
        \multirow{2}{*}{\textbf{Method}} & \textbf{Theoretical Estimation} & \textbf{Experiment (Video-MME)} \\
        ~ & 1 video \& 1 query (30min) &  \makecell{10 video \& 30 queries (avg 41.3min)}\\
        \midrule
        DVD~\cite{zhang2025deep} &  3.98M    &  64.21M  \\
        \textbf{Ours} &  0.08M, \textcolor{teal}{\textbf{1/50 of DVD}} &  1.89M, \textcolor{teal}{\textbf{1/34 of DVD}}\\ 
        \bottomrule
        \end{tabular}
    }
    \vspace{-0.6em}
    \caption{Comparison of token consumption from both theoretical estimation and experimental perspectives.}
    \vspace{-1.0em}
    \label{tab:token_cmp}
\end{minipage}
\end{table}

\subsection{Token Efficiency Analysis}
\label{sec:token}
To quantitatively demonstrate the computational efficiency of our framework, we compare the total token consumption for processing a long-form video from both theoretical estimation and experimental perspectives between the retrieval-centric method DVD~\cite{zhang2025deep} and our proposed VideoARM.

\noindent\textbf{\textit{Theoretical Estimation}.}
DVD pre-segments each video into 10-second clips and performs exhaustive captioning and embedding extraction for all segments before reasoning. Following the official DVD configuration, each clip is sampled at $r_s{=}2$\,FPS and each frame is encoded by GPT-4.1 vision at six $512{\times}512$ patches. According to OpenAI’s tokenization specification for gpt-4.1 vision, this process incurs a base cost of approximately  $t_f{=}1{,}105$ tokens per frame.\footnote{See OpenAI documentation on gpt-4.1 vision tokenization: six 512$\times$512 tiles contribute $\sim$170$\times$6 tokens plus a base cost of $\sim$85 tokens.}
It is important to note that \textit{this estimation only accounts for the visual token consumption} of DVD. In practice, the overall cost of DVD is substantially higher because it also includes tokens for caption generation, text embeddings, and retrieval reasoning,
all of which are omitted here for a fair and conservative comparison with our method.
Let $T_v$ denote the total video duration (in seconds), $r_s$ the frame sampling rate, and $t_f$ the average tokens per frame. The total visual token cost of DVD can be expressed as:
\begin{equation}
C_{\text{DVD}} = T_v \times r_s \times t_f.
\end{equation}
For a 30-minute ($T_v{=}1{,}800$\,s) video, with $r_s{=}2$\,FPS and $t_f{=}1{,}105$, DVD consumes at least \textit{3.98M visual tokens}, excluding additional tokens in captioning and retrieval, representing a conservative lower bound.

Our VideoARM, instead of densely precomputing features, performs agentic video reasoning in an end-to-end manner. Each reasoning iteration consumes approximately $8{,}000$ tokens, including visual, textual, and memory context, and the controller is allowed up to $N$=10 iterations. The total token cost is therefore:
\begin{equation}
C_{\text{ARM}} \leq N \times 8{,}000 = 80{,}000.
\end{equation}
Therefore, VideoARM consumes $\leq$1/50 of the tokens required by DVD for a 30-minute video with one query.

\noindent\textbf{\textit{Experiments on Video-MME}.}
Table \ref{tab:token_cmp} presents the comparison of token consumption on Video-MME. Our VideoARM consumes only 1/34 of the tokens required by DVD when processing 10 videos (average 41.3 minutes) with 30 queries.
This reduction further demonstrates the efficiency of our hierarchical memory and adaptive reasoning design, which eliminates redundant information and significantly reduces token costs for video understanding.

\section{Conclusion}
We introduce VideoARM, a coarse-to-fine video reasoning paradigm over hierarchical multimodal memory (HM$^3$) for long-form video understanding. By leveraging the proposed complementary toolsets and HM$^3$, our method can progressively localize, interpret, and abstract evidence in an adaptive observe–think–act–memorize loop. 
Extensive experiments on prevalent long-video understanding benchmarks demonstrate that VideoARM maintains strong performance while significantly reducing token consumption.

\subsection*{Acknowledgments}
This work was supported by the Zhejiang Provincial Natural Science Foundation of China under Grant No. LQN26F020053, in part by the National Natural Science Foundation of China under Grants No. 62422204, in part by the Fundamental Research Funds for the Provincial Universities of Zhejiang under Grant No. GK259909299001-040, in part by the Zhejiang Provincial Natural Science Foundation of China under Grant No. LRG26F020001, in part by the Scientific Research Innovation Capability Support Project for Young Faculty, in part by the Key Research and Development Program of Zhejiang Province No. 2025C01026, in part by the National Natural Science Foundation of China No.62206082, in part by the National Natural Science Foundation of China under Grants No.62502135, in part by the Zhejiang Provincial Natural Science Foundation of China under Grants No. LQN25F030014.

{
    \small
    \bibliographystyle{ieeenat_fullname}
    \bibliography{main}
}


\lstdefinelanguage{json}{
  basicstyle=\ttfamily\footnotesize,
  breaklines=true,
  breakatwhitespace=false,
  showstringspaces=false,
  columns=fullflexible,
  keepspaces=true,
  frame=none
}
\setcounter{page}{1}
\section*{Suppmentary Material}
\setcounter{section}{0}
\renewcommand{\thesection}{\Alph{subsection}}
\renewcommand{\thesubsection}{\Alph{subsection}}

\subsection{Detailed Pipeline of VideoARM}
\label{sec:pipeline}
The detailed reasoning pipeline of VideoARM is illustrated in Algorithm~\ref{alg:reason_pipeline}. The entire process follows a coarse-to-fine, iterative reasoning loop. 
Concretely, we model VideoARM as an agentic policy $\pi_{\theta}$ that, given the video information $Info$ (\textit{i.e.}, video $V$ and query $Q$) and hierarchical memory $M^{(t)}$ = $(M_s^{(t)}, M_r^{(t)}, M_w^{(t)})$, repeatedly (i) produces a reasoning trace $R_t$ and samples a tool action $(A_t, P_t)$ from the action space $\mathcal{A}$:
\begin{equation}
        (A_t, P_t) \sim \pi_{\theta}(\cdot \mid R_t, M^{(t)}), \quad A_t \in \mathcal{A}.
\end{equation}
(ii) invokes the selected tool and its corresponding temporal tool $T_t$ to obtain an observation $O_t$ and updates the hierarchical memory $M^{(t+1)}$: 
\begin{equation}
\begin{aligned}
    & M^{(t)}_s = T_t(M^{(t)}, Info), \\
    & O_t = A_t(P_t; M^{(t)}_s), \\
    & M^{(t+1)} = M^{(t)} \cup (R_t, O_t),
\end{aligned}
\end{equation}
and (iii) terminates and outputs the final answer $\mathrm{Ans}$ either when the sampled action is $\textsc{Answer}$ or when the number of loops reaches $N$.

\subsection{Subset Analysis in the Ablation Study}
\begin{figure*}
    \centering
    \includegraphics[width=0.98\linewidth]{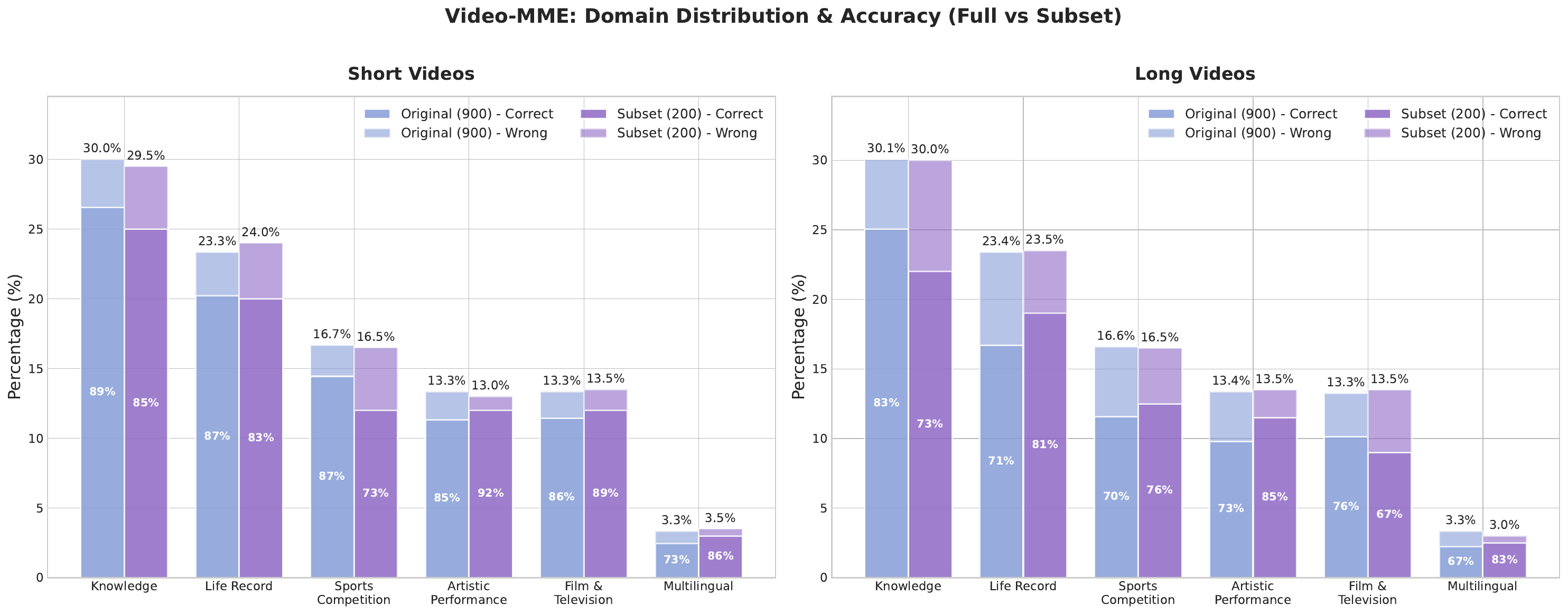}
    \vspace{-0.5em}
    \caption{Comparison of domain distributions and accuracy statistics between the full Video-MME benchmark and the evaluation subset used in our ablation study. The evaluation subset closely matches the domain composition and accuracy trends of the original dataset in both short-video and long-video evaluations.
    }
    \label{fig:framework_supp}
    \vspace{-1.5em}
\end{figure*}
In the ablation study, we construct 200-sample evaluation subsets from the original datasets (\textit{i.e.}, VideoMME-Long) based on a stratified sampling strategy that preserves the original distribution of video types and task categories. Because the domain–task distribution in the full dataset is highly imbalanced, these two factors are strongly coupled, and we therefore treat the domain and task annotations as discrete variables defining a joint empirical distribution.

To approximate this joint distribution when selecting a reduced subset, we 
apply a two-stage stratified sampling strategy. We first compute the empirical 
joint mass function $p(d,t)$ over all domain--task pairs and allocate a target 
sample count $n(d,t)$ to each cell proportional to $p(d,t)$, subject to the 
global 200-sample budget. Domains are given priority: domain-level proportions 
are matched exactly, and the task-level allocation is performed within each 
domain according to their relative frequencies. For cells with extremely small 
mass, we use floor-based rounding with a global residual redistribution step to 
ensure full coverage while avoiding over-sampling rare categories.

Within each $(d,t)$ stratum, we then randomly sample the required number of 
videos without replacement. A final lightweight balancing pass adjusts only 
1--2 samples per subset to eliminate minor rounding discrepancies and ensure 
exact alignment with the target domain--task histogram.

Despite its simplicity, this stratified sampling procedure yields 200-sample 
subsets whose domain closely match those of the 
full dataset, as shown in Figure~\ref{fig:framework_supp}.

\begin{algorithm}[t]
    \caption{Reasoning Pipeline of VideoARM. }
    \SetKwInOut{Input}{Input}
    \SetKwInOut{Output}{Output}
    \Input{Video $V$, Query $Q$, step budget $N$, Controller $C$, Temporal Scoping tools $\mathcal{T}_{ts}$, Multimodal Understanding tools $\mathcal{T}_{mu}$, Long-term perception pool $P_l$, Short-term perception pool $P_s$, Result Memory $M_r$, Working Memory $M_w$, action space $\mathcal{A}=\mathcal{T}_{mu} \cup \{\textsc{Answer}\}$, agentic policy $\pi_{\theta}$}
    \Output{Answer to $Q$}
    \tcp{\textcolor{teal}{\textbf{Initialize VideoARM}}}
    $Info = (V, Q); \quad P^{(0)}_l \leftarrow \text{Sample}(V);$ \\
    $P^{(0)}_s, M^{(0)}_r, M^{(0)}_w \leftarrow \emptyset; \quad M^{(0)}_s = (P^{(0)}_l, P^{(0)}_s);$ \\
    $M^{(0)} = (M^{(0)}_s, M^{(0)}_r, M^{(0)}_w).$ \\
    \For{$t \leftarrow 1$ \KwTo $N$}{
        \tcp{\textcolor{teal}{\textbf{Controller}}}
        Refresh the context of $C$. \\
        $R_t \leftarrow C.reason(M^{(t)}, Info)$ \\
        $(A_t, P_t) \sim \pi_{\theta}(\cdot \mid R_t, M^{(t)}), \quad A_t \in \mathcal{A}$ \\
        \If{$A_t = Answer$}{ 
            \textbf{break}
        }
        \tcp{\textcolor{teal}{\textbf{Temporal Scoping Tools}}}
        Find $T_t \in \mathcal{T}_{ts}$ corresponding to $A_t \in \mathcal{T}_{mu}$. \\
        $M^{(t)}_s \leftarrow T_t(M^{(t)}, Info)$ \\
        \tcp{\textcolor{teal}{\textbf{MM Understanding Tools}}}
        $O_t \leftarrow A_t(P_t; M^{(t)}_s)$ \\
        $M^{(t+1)}_r \leftarrow M^{(t)}_r \cup O_t$  \\ 
        $M^{(t+1)}_w \leftarrow M^{(t)}_w \cup R_t$  \\ 
        $P_s^{(t+1)} \leftarrow \emptyset$ \\
    }
    $\mathrm{Ans} \leftarrow C.answer(M^{(t^*)}, Info).$ \\
    \Return{$\mathrm{Ans}$}
\label{alg:reason_pipeline}
\end{algorithm}

\subsection{Limitation and Discussion}
VideoARM already achieves strong performance on long-form video understanding while effectively reducing token consumption. However, the current implementation of VideoARM has several concrete limitations:

First, VideoARM is practically limited by its dependence on the quality of the initial temporal and visual sampling, which seeds the subsequent reasoning chain. The controller builds early hypotheses from sparsely sampled frames and short clips; if these miss brief but crucial moments or extremely fine-grained details (\textit{e.g.}, a small object, a subtle gesture, or a one-frame change), the agent may struggle to initiate a valid reasoning trajectory. 

Although our \textit{Temporal Scoping} tools can sometimes recover missed evidence by sampling additional clips and revisiting nearby regions, this process is neither guaranteed nor efficient, as there is no explicit mechanism to detect early mis-localization and trigger targeted re-exploration. Consequently, the performance on very fine-grained or fleeting events remains constrained by this initial sampling bottleneck.

\begin{figure}
    \centering
    \includegraphics[width=0.99\linewidth]{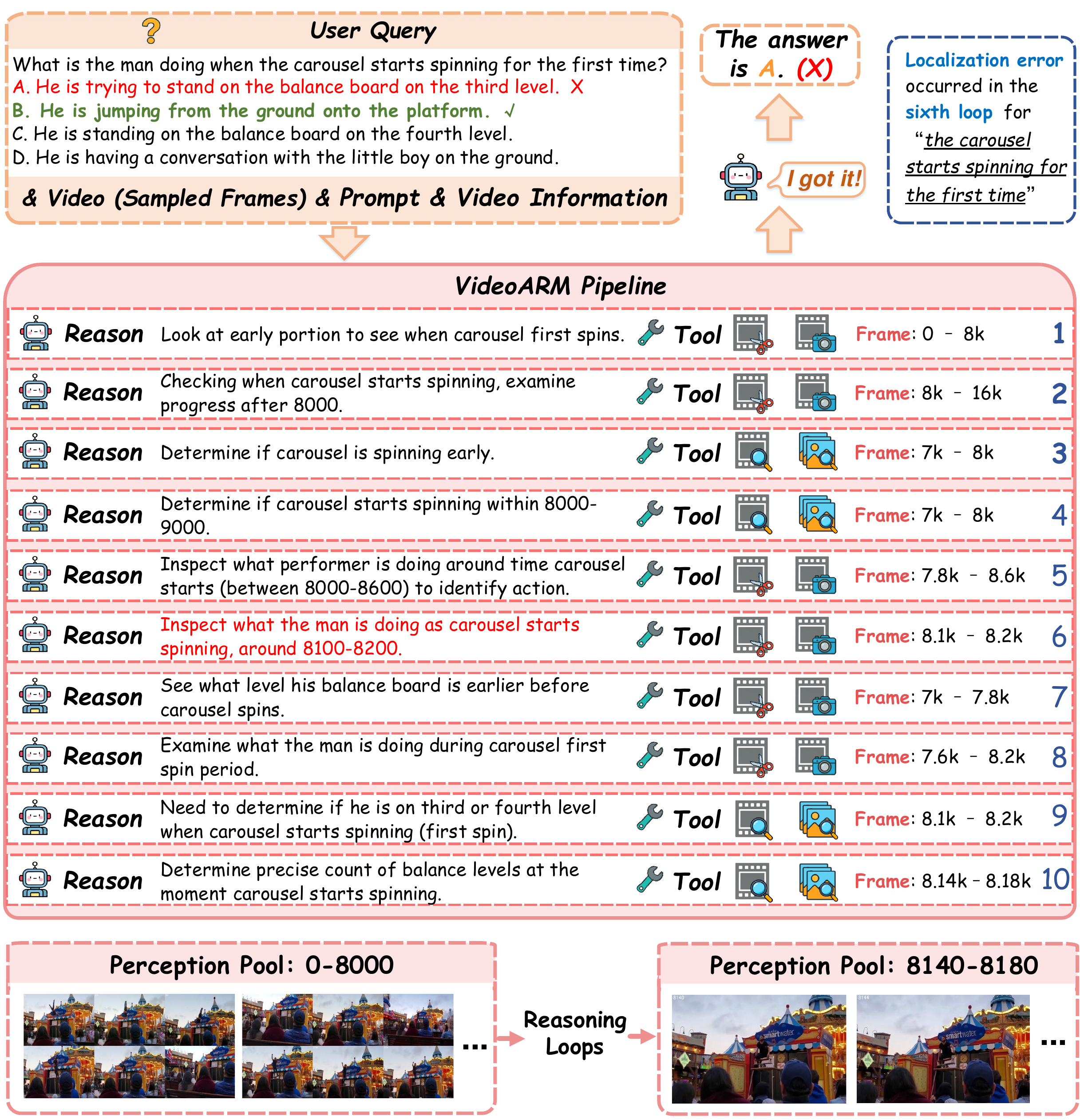}
    \vspace{-0.6em}
    \caption{Visualization of the failure case of VideoARM pipeline. 
    }
    \label{fig:fail}
    \vspace{-1.5em}
\end{figure}

As shown in Figure~\ref{fig:fail}, 
when the object “the carousel” is heavily occluded and the query-relevant event occurs within a very short time interval, VideoARM struggles to localize the most relevant frames, rendering subsequent correct reasoning ineffective.

Second, the current instantiation of VideoARM still relies heavily on proprietary, closed-source MLLMs for both the controller and worker agents, and our attempts to migrate the framework to fully open-source backbones remain unsatisfactory. In our experiments, directly replacing closed-source models with state-of-the-art open-source alternatives causes a clear performance drop, largely because many open-source models still exhibit weaker multi-step reasoning and long-context planning, and struggle to align fine-grained visual evidence under VideoARM’s “jigsaw-style” video interface, where temporally distant frames and clips are densely arranged in grids. Nevertheless, we also see encouraging signs that this gap is narrowing—for example, in our tests, the latest open-source model Qwen3-VL already matches or even surpasses our GPT-4o-based configuration on several challenging benchmarks—suggesting that a fully open and self-hostable version of VideoARM is a realistic and promising direction as open-source reasoning models continue to improve.

\begin{figure*}
    \centering
    \includegraphics[width=0.98\linewidth]{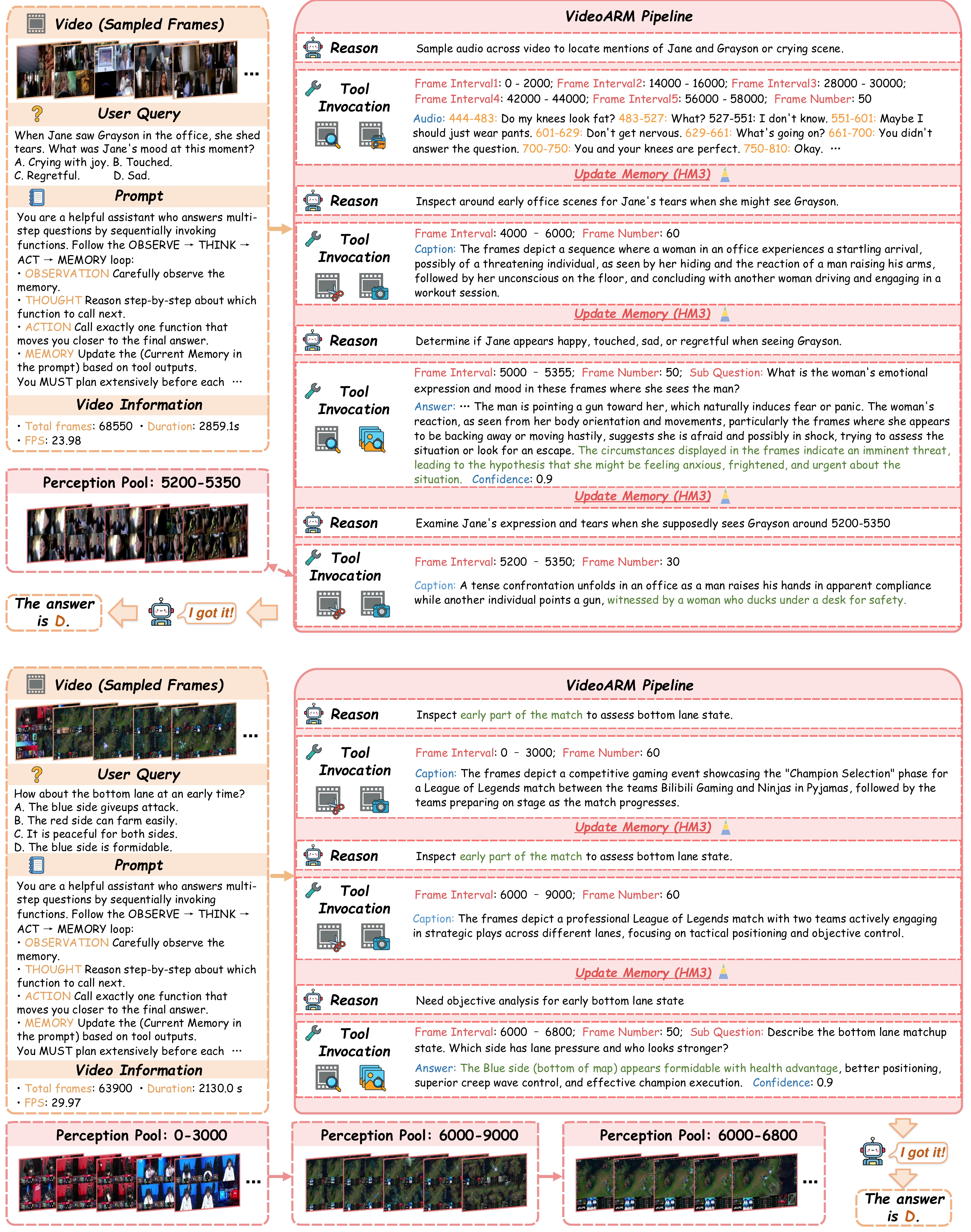}
    \vspace{-0.3em}

    \caption{Additional visualizations of the VideoARM pipeline. For brevity, the \textit{perception pool} displays only the \textit{short-term} or \textit{long-term} pool invoked by the tool in the current loop.
    }
    \label{fig:pipeline_more}
    \vspace{-0.6em}
\end{figure*}
\subsection{More Visualization Results}
Figure~\ref{fig:pipeline_more} shows the additional visualizations of our VideoARM pipeline. Our method progressively identifies key temporal intervals and retrieves relevant information through the autonomous invocation of tools, ultimately providing the correct answer.

\subsection{Prompts of VideoARM}
\subsubsection{Agent System Prompt}
\begin{table*}[ht]
\centering
\footnotesize
\begin{tabular}{p{3.1cm} p{9.3cm} p{3.2cm}}
\toprule
\textbf{Prompt Name} & \textbf{Purpose} & \textbf{Used By} \\
\midrule
Agent System Prompt 
& Defines global reasoning rules (OBSERVE–THINK–ACT–MEMORY), tool-calling constraints, and planning behavior 
& Controller \\

Agent User Prompt
& Formats the query and memory into a structured prompt for each reasoning step 
& Controller \\

Tools List 
& Defines available tools, their signatures, and call behaviors for the agent system 
& Controller \\

Scene Caption Prompt 
& Produces concise event or appearance descriptions for selected clips 
& Scene Snapper \\

Clip Analyzer Tool Prompt 
& Inspect selected frame intervals and return analysis or evidence 
& Clip Analyzer  \\

\bottomrule
\end{tabular}
\vspace{-0.2em}
\caption{Summary of all prompts used in VideoARM. Each prompt corresponds to one stage of the reasoning pipeline.}
\label{tab:prompt_summary}
\vspace{-0.5em}
\end{table*}

Based on the categorization in Table \ref{tab:prompt_summary}, we next present the full text of each prompt in a modular and self-contained manner. For clarity and reproducibility, the following subsections list the exact system prompt, user prompt, tools list and tool prompts used in VideoARM.

\subsubsection{Agent System Prompt}
\begin{tcolorbox}[
    enhanced,
    colback=gray!3,
    colframe=black!25,
    boxrule=0.25pt,
    arc=2pt,
    left=5pt,
    right=5pt,
    top=5pt,
    bottom=5pt,
    width=\linewidth
]
\small  
You are a helpful assistant who answers multi-step questions by sequentially invoking functions. 
Follow the \textbf{OBSERVE → THINK → ACT → MEMORY} loop:
\begin{itemize}[leftmargin=1em]
    \item \textbf{OBSERVATION} — Carefully observe the memory.
    \item \textbf{THOUGHT} — Reason step-by-step about which function to call next.
    \item \textbf{ACTION} — Call exactly one function that moves you closer to the final answer.
    \item \textbf{MEMORY} — Update the current memory based on tool outputs.
\end{itemize}

\textbf{Important:} You MUST plan extensively before each function call, and reflect on the outcomes of previous calls.  
If uncertain about code structure or video content, \textbf{use tools to inspect rather than guessing}.  
Do not rely on blind function calls — this degrades reasoning quality.

Each extracted frame contains the global frame index in white text.  
Each image you see is a \textbf{3×2 mosaic}.  
Give the final answer only when you are confident; otherwise continue tool calls.

\end{tcolorbox}
In our design, the agent is instantiated as a multi-step reasoning assistant that operates under a clearly defined control regime. The system prompt formalizes this behavior by enforcing an explicit OBSERVE → THINK → ACT → MEMORY loop, which structures every reasoning round into four interpretable and verifiable stages. This design ensures that each action is grounded in the agent’s current evidence, and that all new information is immediately incorporated into the hierarchical memory for subsequent steps. To maintain stable reasoning behavior, the prompt further imposes several core rules: the agent must plan before each tool call, reflect on the returned result, inspect code or video segments via tools whenever uncertainty arises, and avoid blind or unmotivated tool invocations. We additionally define visual conventions—global frame indices overlaid in white and all inspected images formatted as 3×2 mosaics—to standardize the perceptual input seen by the controller. Finally, the agent is instructed to produce a final answer only when it has accumulated sufficient evidence to be confident, ensuring that our design favors precision and robustness in long-form video reasoning.

\subsubsection{Agent User Prompt}
\begin{tcolorbox}[
    enhanced,
    colback=gray!3,
    colframe=black!25,
    boxrule=0.25pt,
    arc=2pt,
    left=5pt,
    right=5pt,
    top=5pt,
    bottom=5pt,
    width=\linewidth
]
\small
\textbf{Video Information}
\begin{itemize}[leftmargin=1em]
    \item \textbf{Total frames}: \textcolor{blue!60!black}{\{}\textcolor{orange}{total\_frames}\textcolor{blue!60!black}{\}}
        \item \textbf{Duration}: \textcolor{blue!60!black}{\{}\textcolor{orange}{duration:.1f}\textcolor{blue!60!black}{\}} seconds
    \item \textbf{FPS}: \textcolor{blue!60!black}{\{}\textcolor{orange}{fps:.2f}\textcolor{blue!60!black}{\}}
\end{itemize}

\textbf{Current Memory (JSON)}:\\
\textcolor{blue!60!black}{\{}\textcolor{orange}{memory\_json}\textcolor{blue!60!black}{\}}\\
\textbf{Question}:

\textcolor{blue!60!black}{\{}\textcolor{orange}{question\_text}\textcolor{blue!60!black}{\}}\\
\textbf{Options:}

\textcolor{blue!60!black}{\{}\textcolor{orange}{question\_text\_with\_options}\textcolor{blue!60!black}{\}}\\
Respond with only the letter (A, B, C, or D) of the correct option.
\end{tcolorbox}

This user prompt template provides a standardized interface between the controller and the underlying tools. Rather than issuing free-form queries, the controller is required to ground each step in the current question, video metadata, and memory state, and to explicitly state missing information before selecting a tool. This structuring encourages the agent to decompose the problem into a sequence of targeted evidence-gathering operations, instead of jumping directly to an answer. By forcing the controller to justify every tool call and to specify concrete inputs such as frame intervals or clips, the template makes the reasoning process more interpretable and reduces spurious or redundant tool invocations, which is crucial for stable long-form video reasoning.

\begin{tcolorbox}[
  enhanced,
  breakable,
  colback=codebg,
  colframe=black!20,
  boxrule=0.3pt,
  arc=2pt,
  left=4pt,
  right=4pt,
  top=4pt,
  bottom=4pt
]
\begin{Verbatim}[fontsize=\footnotesize, breaklines=true, breakanywhere=true]
[
{
    "type": "function",
    "function": {
      "name": "scene_snapper",
      "description": "Scene Snapper - Navigate to frame ranges, extract images (default 30 frames), and return an auto-generated caption.",
      "parameters": {
        "type": "object",
        "properties": {
          "frame_ranges": {
            "type": "array",
            "items": {
              "type": "object",
              "properties": {
                "start_frame": { "type": "integer", "description": "Starting frame number" },
                "end_frame": { "type": "integer", "description": "Ending frame number" }
              },
              "required": ["start_frame", "end_frame"]
            },
            "description": "Array of frame ranges to view. Frames are distributed proportionally across ranges."
          },
          "num_frames": {
            "type": "integer",
            "description": "Number of frames to extract across ranges: 30/60/90/150 (default: 30).",
            "enum": [30, 60, 90, 150],
            "default": 30
          },
          "reason": { "type": "string", "description": "Short rationale describing why the tool is being invoked." }
        },
        "required": ["frame_ranges", "reason"]
      }
    }
  },
  {
    "type": "function",
    "function": {
      "name": "audio_transcripter",
      "description": "Extract and transcribe audio from specific frame ranges using whisper-1 API. Returns only segments with global frame indices and text.",
      "parameters": {
        "type": "object",
        "properties": {
          "frame_ranges": {
            "type": "array",
            "items": {
              "type": "object",
              "properties": {
                "start_frame": { "type": "integer", "description": "Starting frame number" },
                "end_frame": { "type": "integer", "description": "Ending frame number" }
              },
              "required": ["start_frame", "end_frame"]
            }
          },
          "reason": { "type": "string", "description": "Short rationale describing why the tool is being invoked." }
        },
        "required": ["frame_ranges", "reason"]
      }
    }
  },
  {
    "type": "function",
    "function": {
      "name": "clip_analyzer",
      "description": "Analyze frames within a specific range by asking a question.",
      "parameters": {
        "type": "object",
        "properties": {
          "frame_range": {
            "type": "object",
            "properties": {
              "start_frame": { "type": "integer", "description": "Starting frame number" },
              "end_frame": { "type": "integer", "description": "Ending frame number" }
            },
            "required": ["start_frame", "end_frame"]
          },
          "question": { "type": "string", "description": "Question to ask about the frame range content" },
          "reason": { "type": "string", "description": "Short rationale describing why the tool is being invoked." }
        },
        "required": ["frame_range", "question", "reason"]
      }
    }
  }
]
\end{Verbatim}
\end{tcolorbox}

The tools list prompt serves as a formal specification for all tool interfaces that the agent may invoke during reasoning. By presenting tools in a structured, argument-explicit format, we ensure that the controller understands both the capabilities and the constraints associated with each tool. This prevents misuse, reduces invalid calls, and promotes interpretability by making each step of the reasoning pipeline verifiable.
\subsubsection{Scene Snapper Prompt}
\begin{tcolorbox}[
    colback=gray!3,
    colframe=black!25,
    boxrule=0.25pt,
    arc=2pt,
    left=5pt,
    right=5pt,
    top=5pt,
    bottom=5pt,
    width=\linewidth
]
\small

You are a video frame captioning assistant. Each extracted frame displays the global frame index in white text at the top-left. Each picture you see contains one 3×2 mosaic (a single image composed of 6 frames). row-major from top-left to bottom-right

Caption the provided\textcolor{blue!60!black}{\{}\textcolor{blue}{len(}\textcolor{orange}{frame\_paths}\textcolor{blue}{)}\textcolor{blue!60!black}{\}} frames sampled uniformly from frame range \textcolor{blue!60!black}{\{}\textcolor{orange}{start\_frame}\textcolor{blue!60!black}{\}} -\textcolor{blue!60!black}{\{}\textcolor{orange}{end\_frame}\textcolor{blue!60!black}{\}}. The frames represent a continuous sequence from the video. Describe the main scene or action in these frames using a concise English sentence.

\end{tcolorbox}

The Scene Caption Prompt defines a lightweight abstraction mechanism that provides scene-level summaries for longer clips. While the Clip Analyzer operates on fine-grained evidence, the Scene Captioner complements it by generating high-level semantic context that helps the agent track global events, locations, or transitions. 

\subsubsection{Clip Analyzer Prompt}

\begin{tcolorbox}[
    colback=gray!3,
    colframe=black!25,
    boxrule=0.25pt,
    arc=2pt,
    left=5pt,
    right=5pt,
    top=5pt,
    bottom=5pt,
    width=\linewidth
]
\small

    You are an expert video frame analyst. Analyze the provided \textcolor{blue!60!black}{\{}\textcolor{blue}{len(}\textcolor{orange}{frame\_paths}\textcolor{blue}{)}\textcolor{blue!60!black}{\}} frames sampled uniformly from frame range \textcolor{blue!60!black}{\{}\textcolor{orange}{start\_frame}\textcolor{blue!60!black}{\}} -\textcolor{blue!60!black}{\{}\textcolor{orange}{end\_frame}\textcolor{blue!60!black}{\}}and answer the given question. The frames represent a continuous sequence from the video. Analyze them collectively to provide a comprehensive answer. Each extracted frame displays the global frame index in white text at the top-left.

Analyze this frame sequence and answer the question below.
IMPORTANT: Your response must include a confidence score between 0.0 and 1.0 indicating how confident you are in your answer.
Format your response as:\\
Answer: [Your detailed answer here]\\
Confidence: [0.0-1.0]\\
\textbf{Question}:\\
\textcolor{blue!60!black}{\{}\textcolor{orange}{question\_text}\textcolor{blue!60!black}{\}}\\
\textbf{Options:}

\textcolor{blue!60!black}{\{}\textcolor{orange}{question\_text\_with\_options}\textcolor{blue!60!black}{\}}\\
\end{tcolorbox}

The Clip Analyzer Prompt specifies the behavior of VideoARM’s fine-grained visual inspection tool. The prompt ensures that the model focuses on extracting concrete evidence—actions, object states, and temporal transitions—rather than prematurely interpreting or answering the user’s question. By explicitly specifying the interval and requiring uniformly sampled frames, we enforce consistency across tool calls and prevent biases caused by arbitrary sampling. The structured output schema further improves interpretability, allowing the controller to reason over analyzable evidence rather than free-form text. This modular design is crucial for long-form videos, where localized events must be inspected precisely and repeatedly across multiple reasoning rounds.

\end{document}